%% file: top.tex
\documentclass{article}

\usepackage[final]{corl_2018} 
\usepackage[utf8]{inputenc}
\input{shortcuts.tex}

\usepackage[resetlabels]{multibib}
\newcites{New}{References}

\title{Conditional Affordance Learning\\for Driving in Urban Environments}

\author{
Axel Sauer\textsuperscript{\normalfont{1,2}} \quad
Nikolay Savinov\textsuperscript{\normalfont{1}} \quad
Andreas Geiger\textsuperscript{\normalfont{1,3}}\\[2mm]
\textsuperscript{1} Computer Vision and Geometry Group, ETH Z{\"u}rich \\
\textsuperscript{2} Chair of Robotics Science and System Intelligence, Technical University of Munich \\
\textsuperscript{3} Autonomous Vision Group, MPI for Intelligent Systems and University of T{\"u}bingen
}
\begin{document}
\maketitle

\vspace{-0.65cm}
\begin{abstract}
Most existing approaches to autonomous driving fall into one of two categories:
modular pipelines, that build an extensive model of the environment, and imitation learning approaches, that map images directly to control outputs. A recently proposed third paradigm, direct perception, aims to combine the advantages of both by using a neural network to learn appropriate low-dimensional intermediate representations. However, existing direct perception approaches are restricted to simple highway situations, lacking the ability to navigate intersections, stop at traffic lights or respect speed limits. In this work, we propose a direct perception approach which maps video input to intermediate representations suitable for autonomous navigation in complex urban environments given high-level directional inputs. Compared to state-of-the-art reinforcement and conditional imitation learning approaches, we achieve an improvement of up to 68 \% in goal-directed navigation on the challenging CARLA simulation benchmark. In addition, our approach is the first to handle traffic lights and speed signs by using image-level labels only, as well as smooth car-following, resulting in a significant reduction of traffic accidents in simulation.
\end{abstract}


\input{tex/01_introduction.tex}
\input{tex/02_relatedworks.tex}
\input{tex/03_CAL.tex}
\input{tex/04_results.tex}
\input{tex/05_conclusion.tex}

\newpage
\input{tex/A_00.tex}
\clearpage

\acknowledgments{We would like to thank Vanessa Diedrichs for many helpful discussions about autonomous driving and data visualization and for her general support.}

\bibliography{bibliography_long,bibliography,bibliography_axel}  

\end{document}

%% file: tex/01_introduction.tex
\section{Introduction}\label{sec:introduction}

An autonomous vehicle is a cognitive system and hence follows the concept of a rational agent: in order to operate safely it must accurately observe its environment, make robust decisions and perform actions based on these decisions \citep{russell2016artificial}. Autonomous navigation can be described as a mapping function from sensory input to control output. To implement this function, three major approaches have been proposed: modular pipelines, imitation learning, and direct perception.

Most autonomous vehicles are based on \textit{modular pipelines} (MP) \citep{thrun_stanley:_2006,ziegler_making_2014}. An MP splits the autonomous driving problem into smaller, easier subproblems: perception, path planning, and control. The approach often relies on various sensors in order to produce a consistent representation of the surrounding environment. A driving decision is then made based on this representation.
While MPs are relatively interpretable due to their modularity, they rely on complex intermediate representations which are manually chosen (\eg, optical flow), often hard to estimate with sufficient accuracy and might not necessarily be the optimal choice for solving the sensorimotor control task. 
Furthermore, MPs require large amounts of annotated data which can be costly to obtain, \eg, pixel-wise semantic segmentation for training neural networks or high-definition maps for localization.

The second approach, \textit{imitation learning} (IL) \citep{pomerleau_alvinn:_1989}, maps directly from raw input to control output and is learned from data in an end-to-end fashion, skipping the intermediate step of building an explicit environment model.
In contrast to MPs, data for training IL-based navigation systems can be collected with relative ease, \ie, by driving around and recording the video footage together with the driven trajectory.
However, IL approaches are faced with the problem of learning a very complex mapping (from raw input to control) in a single step. Thus, a model with high capacity as well as a very large amount of training data is required for handling the large variety of real world situations which may be encountered at test time.
Besides, IL lacks transparency as it is hard to comprehend the internal decision-making process in a neural network. This raises security concerns: a driving system that does not act transparently might not be trusted or  confidently used by humans.

The third approach, \textit{direct perception} (DP) \citep{chen_deepdriving:_2015}, aims to combine the advantages of MP and IL. Instead of predicting a detailed representation of the environment, the goal in DP is to predict a low-dimensional intermediate representation of the environment which is then used in a conventional control algorithm to maneuver the vehicle.
Thus, DP neither requires the network to learn the complex sensorimotor control problem end-to-end, nor does it assume the availability of datasets with pixel-level or box-level labels that are significantly more time-consuming to obtain than image-level labels.
\citet{chen_deepdriving:_2015} demonstrated good results when applying this approach to highway driving using an open-source simulator. However, highway driving is a rather easy task compared to driving in rural or urban areas. When considering navigation in urban areas, several difficulties are added: the agent must obey traffic rules (speed limits, red lights, etc.), take into account possible obstacles on the road (\eg, pedestrians crossing the street), and handle  junctions with more than one possible direction.

At the core of the direct perception approach lies the choice of intermediate representation to be predicted. Ideally, this representation should be of low dimensionality while comprising all the necessary information for making a driving decision. One choice of such representation are \textit{affordances}, attributes of the environment which limit the space of allowed actions. As an example, consider the distance to the vehicle ahead which limits the ability of the ego-vehicle to speed up.
Affordances can also be conditioned on high-level driving decisions as provided, \eg, by a conventional navigation system.
Consider a vehicle at an intersection: the distance to the centerline changes with the desired turn we would like the vehicle to take (and thus the lane we would like it to follow).

\begin{figure}
  \centering
  \includegraphics[width=0.95\linewidth]{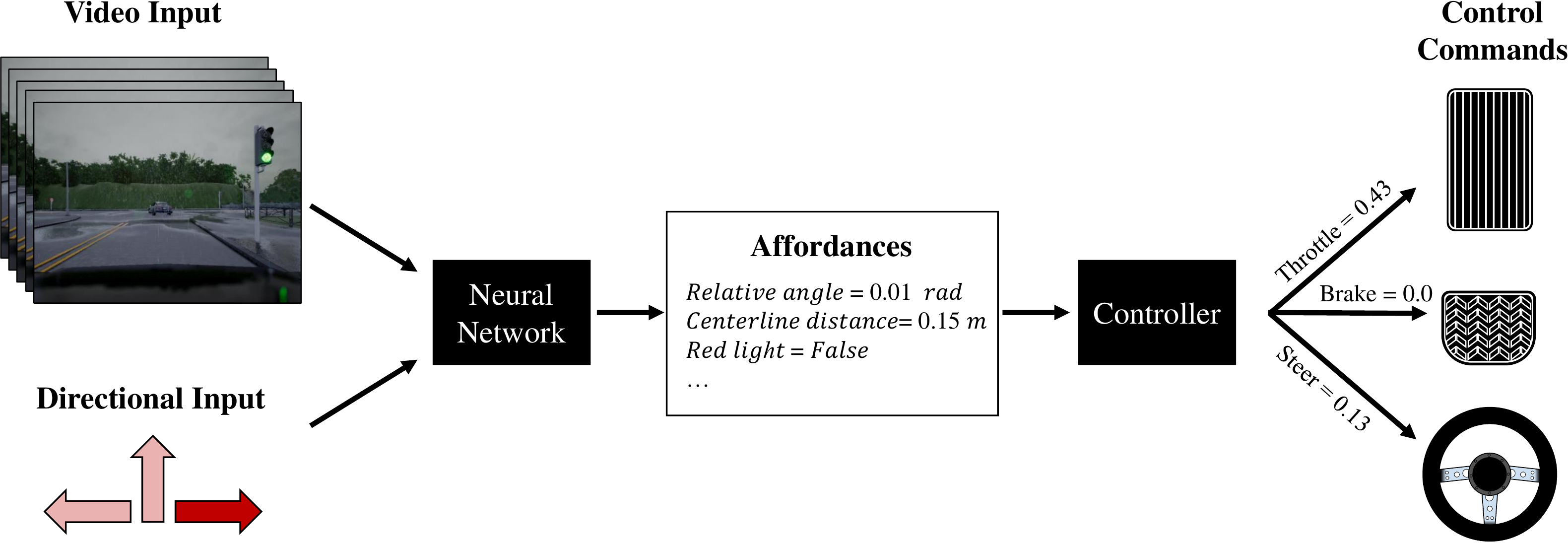}
  \captionof{figure}{{\bf Conditional Affordance Learning (CAL) for Autonomous Urban Driving.} The input video and the high-level directional input are fed into a neural network which predicts a set of affordances. These affordances are used by a controller to calculate the control commands.}
  \label{fig:pull_figure}  
  \vspace{-0.5cm}
\end{figure}

The goal of this work is to generalize the direct perception approach to the urban setting. We develop novel intermediate representations in the form of affordances suitable for urban navigation as well as conditional models where the decision is not based on the image input alone, but also on high-level navigation commands as provided by a consumer grade navigation system (\eg, ``turn left at the next intersection''). Moreover, our system is designed to drive more safely and comfortably, alleviating the common problem of jerky driving behavior in existing approaches for autonomous driving \cite{metz_nauseating_2018}. We coin the developed approach ``Conditional Affordance Learning'' (CAL) as the predicted affordances and the model output are conditioned on high-level navigation  commands. Figure \ref{fig:pull_figure} shows an overview of our approach. The main contributions of our work are summarized as follows:

\begin{itemize}[leftmargin=*]
\item We develop novel intermediate representations which are low-dimensional, yet provide the necessary information to allow for driving in urban environments.
\item We deploy a conditional navigation model which allows for taking high-level directional commands prior to intersections. We further implement a control algorithm which allows for a smooth ride while obeying traffic rules.
\item We provide an in-depth evaluation of different network architectures and parameterizations. Our network is able to predict all intermediate representations in a single forward pass.
\item We demonstrate the benefits of video-based representation learning for navigation.  Prior works either operate frame-by-frame or do not evaluate in an online setting. 
\end{itemize}

The code and our trained models can be found at \url{https://github.com/xl-sr/CAL}.

%% file: tex/02_relatedworks.tex
\section{Related Work}\label{sec:relatedwork}
Modular pipelines are the most popular approach to autonomous driving and are employed by most research projects \cite{thrun_stanley:_2006,ziegler_making_2014}. To build the environmental model, the perception stack needs to detect all aspects of the traffic scene that are likely to be relevant for the driving decision. These detection tasks are usually trained and solved separately \cite{geiger_vision_2013}, most recently by exploiting deep neural networks, \eg, for object detection \citep{redmon2016you,ren2015faster}, image segmentation \cite{he2017mask,li2017fully}, or motion estimation \citep{sun2017pwc,guney2016deep}. This information can be aggregated into a model of the environment \cite{zhang_understanding_2013, geiger_3d_2014} and a planning module generates an obstacle-free trajectory which is executed by the control module \citep{schwarting2018planning}.

The first implementation of the imitation learning (IL) approach in the real world, ALVINN \cite{pomerleau_alvinn:_1989}, dates back to the 80s. More recently, \citet{muller_off-road_2006} employ a similar system to navigate an off-road truck in a variety of terrains, weather, and lighting conditions while avoiding obstacles. Using more computational power and a modern CNN architecture, contemporary approaches such as the one by \citet{bojarski_end_2016} demonstrate impressive performance on real-world tasks like highway following and driving in flat courses.
\citet{codevilla_end--end_2017} propose a conditional imitation learning formulation that is able to traverse intersections based on high-level navigational commands. 

The idea of direct perception is to predict several affordance indicators describing the driving scene. The concept of affordances was originally proposed by \citet{gibson1966senses} in the field of psychology and has been applied to the autonomous driving task by \citet{chen_deepdriving:_2015}, demonstrating strong performance in a racing simulation. \citet{al-qizwini_deep_2017} improve the original approach of \citet{chen_deepdriving:_2015} by analyzing different CNNs for the mapping from image to indicators. They find GoogLeNet \citep{szegedy_going_2015} and VGG16 \citep{simonyan_very_2014} to perform best on this task.

Due to the difficulties of training and testing in real environments, driving simulators are often exploited in related work.
Two popular driving simulators for research are the open source car racing game TORCS \cite{chen_deepdriving:_2015,al-qizwini_deep_2017,huang_evolving_2015} and the commercial game Grand Theft Auto V (GTA V) \cite{richter_playing_2016,ebrahimi_gradient-free_2017}. However, TORCS is neither photorealistic nor complex enough, lacking important scene elements such as intersections, pedestrians and oncoming traffic. GTA V can be considered photorealistic, but is closed source and hence limited in its ability to customize and control the environment. In this work, we use the recently released open-source simulator CARLA \cite{dosovitskiy_carla:_2017} which provides a trade-off between realism and flexibility, thus addressing some of the problems of previous simulators.

%% file: tex/03_CAL.tex
\section{Conditional Affordance Learning}\label{sec:CAL}

\begin{figure}
\centering
  \includegraphics[width=0.84\linewidth]{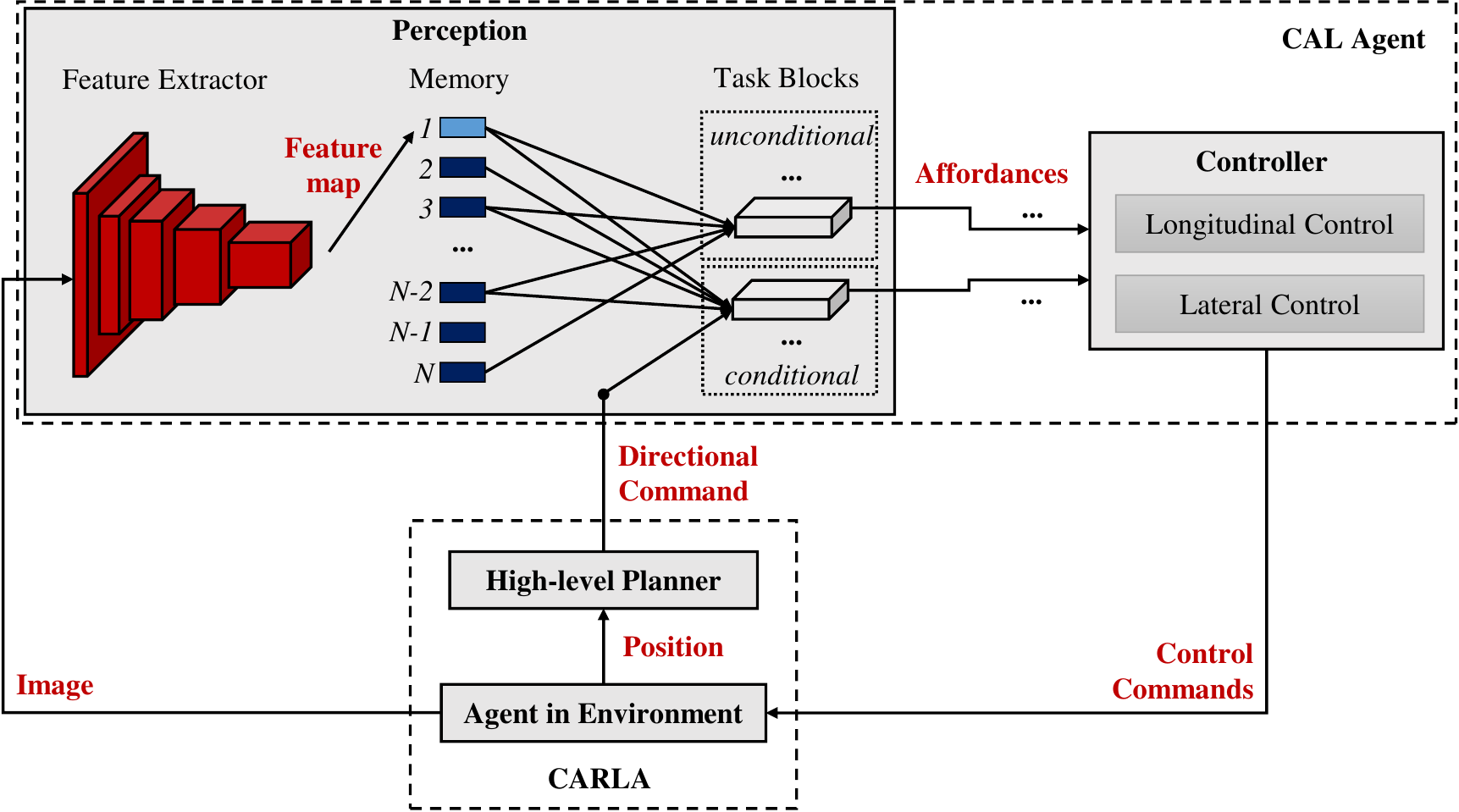}
  \captionof{figure}{{\bf Overview of our System.} The CAL agent (top) receives the current camera image and a directional command (``straight'',``left'',``right'') from CARLA \cite{dosovitskiy_carla:_2017}.
  The feature extractor converts the image into a feature map. The agent stores the last $N$ feature maps in memory, where $N$ is the length of the input sequence required for the perception stack. This sequence of feature maps, together with the directional commands from the planner, are exploited by the task blocks to predict affordances. Different tasks utilize different temporal receptive fields and temporal dilation factors.
   The control commands calculated by the controller are sent back to CARLA which updates the environment and provides the next observation and directional command.}
 \label{fig:system_overview}
 \vspace{-0.5cm}
\end{figure}

Our autonomous driving setup consists of an agent which interacts with the environment as illustrated in Figure~\ref{fig:system_overview}. A high-level command for the maneuver to be performed (e.g., ``go straight'', ``turn left'', ``turn right'') is provided by a planner, mimicking one of the most common driving scenarios: a human driver (the agent) following the commands of a navigation system (the planner).
CARLA uses an A* topological planner to generate these commands.
Based on the high-level command and the observation from the environment, the agent calculates the value of throttle, brake and steering wheel.
Those values are sent back to the environment model which provides the next observation and high-level command, and the process continues until the goal is reached. 

In this work, we focus on visual-based navigation: we only consider observations from a single front-facing camera.
Cameras have the advantage that they are ubiquitous and cheap.
In addition, they provide information that laser scanners or radar sensors cannot provide, \eg, the color of a traffic light or the type of a speed sign.
Thus they are the only sensor that allows to solve the task without any additional sensors.
Note that a monocular setup is comparable to how humans solve the driving task. The stereo baseline of the human visual system is insufficient to perceive accurate depth at distances of 50 meters and beyond based on this cue alone. 

In the following we will discuss the intermediate representations required for autonomous navigation in urban environments and the submodules of the CAL agent for perception and control. 

\subsection{Affordances}

A good intermediate representation should be low-dimensional, yet expressive enough to meet the necessary requirements for autonomous navigation: (i) the agent should be able to drive from A to B as fast as possible while obeying traffic rules. (ii) infractions against traffic rules should be avoided at all times. In this work, we consider $6$ types of infractions: driving on the wrong lane, driving on the sidewalk, running a red light, colliding with other vehicles, hitting pedestrians and hitting static objects. Moreover, the speed limit must be obeyed at all times. (iii) the agent should be able to provide a pleasant driving experience. The car should stay in the center of the road and take turns smoothly. In addition, it should be able to follow a leading car at a safe distance. 

\begin{figure}
\centering    
    \begin{minipage}{0.67\linewidth}
      \centering    
      \resizebox{1.0\linewidth}{!}{
      \ra{2}
      \begin{tabular}{@{}lclcl@{}}
      \toprule
      Type          & Conditional & Affordances  					 & Acronym & Range of values            \\ \midrule
      discrete      & No          & Hazard stop                 	 &	-	   & $\in \{True, False\}$      \\ 
                    &             & Red Traffic Light              &  -       & $\in \{True, False\}$      \\ 
                    &             & Speed Sign {[}km/h{]}          & 	-	   & $\in \{None, 30, 60, 90\}$ \\  \midrule                 
      continuous    & No 			& Distance to vehicle {[}m{]}    & 	$\ell$	   & $\in [0, 50]$ 				\\ \midrule          
      continuous    & Yes         & Relative angle {[}rad{]}       &  $\psi$	   & $\in [-\pi, \pi]$          \\ 
                    &     		& Distance to centerline {[}m{]} &	$d$	   & $\in [-2, 2]$              \\       
      \bottomrule
      \end{tabular}
       }%
  	\end{minipage}%
	\hspace{0.05cm}
    \begin{minipage}{0.32\linewidth} 
    	\includegraphics[width=\textwidth]{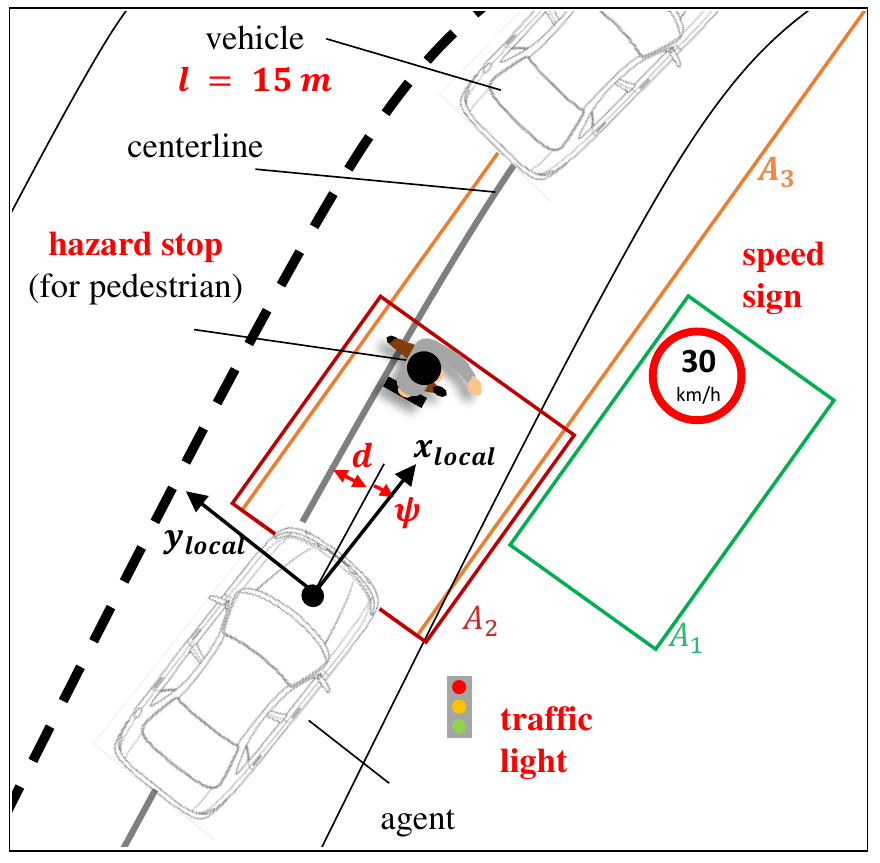}
  	\end{minipage}
	\caption{{\bf Affordances.} Left: We categorize affordances according to their  type (discrete/continuous) and  whether they are conditional (dependent on directional input) or unconditional. Right: Illustration of the affordances (red) and observation areas used by our model.}
	\label{fig:affordances}
	\vspace{-0.5cm}
\end{figure}

Based on these requirements we propose the affordance set shown in Figure~\ref{fig:affordances} (left).
The agent should stop for \textbf{red traffic lights} and change its speed according to the current \textbf{speed sign}. For this purpose we define an \textit{observation area} $A_1$ in the local $(x, y)$ coordinates of the agent, as shown in Figure~\ref{fig:affordances} (right). If a traffic light or a speed sign is within $A_1$, the corresponding affordance indicator switches to \textit{True} or to the speed limit value.
The agent keeps an internal memory of the speed limit. Once a speed sign is passed, the current speed limit is updated and the agent adjusts its speed accordingly.  
If an obstacle is in front of the car --- located within an \textit{observation area} $A_2$ --- the agent should perform a \textbf{hazard stop} --- come to a complete stop as quickly as possible. In CARLA, possible road obstacles are cars and pedestrians.
For driving in the center of the road, the position and the orientation of the car on the road must be known. The position is defined by the \textbf{distance to centerline} $d$, the lateral distance of the center of the vehicle's front axle to the closest point on the centerline. The \textbf{relative angle} $\psi$ describes the orientation of the car with respect to the road tangent. 
To achieve steady vehicle following, the \textbf{distance to the vehicle} $\ell$ ahead must be known.
If a car is located in an \textit{observation area} $A_3$, we measure $\ell$ as the shortest distance between the bounding boxes of both cars, otherwise set $\ell=50$. 

\subsection{Perception}

We formulate perception as a multi-task learning (MTL) problem: using a single neural network we predict all affordances in a single forward pass. 
This yields a computationally efficient solution -- for a forward pass, the system (Network + Controller) takes 50 ms on average on a NVIDIA GTX 1080 Ti GPU. This is well within real-time requirements for autonomous driving systems, where latency should reach levels below 100 ms \citep{lin2018architectural}.
Furthermore, MTL has been shown to learn internal representations which are able to improve the network's generalization abilities \citep{Caruana_multitasklearning_1993}. \citet{xu_end_2016} demonstrate this effect in the context of autonomous driving. In our network, all convolutional layers are shared between tasks. The output of the last convolutional layer is used as input for the task-specific branches which we will call ``task blocks'' in the following.

\textbf{Neural network architecture.} To extract features from raw input images, we use the convolutional layers of a VGG16 network \cite{simonyan_very_2014} pretrained on ImageNet \cite{deng_imagenet_2009}. The simple structure of the network make it an appealing choice for a feature extractor with real-time requirements.  The feature extractor is applied to every image in the input sequence to extract a sequence of feature maps. The task blocks are shallow networks and consist of a task-specific layer, followed by a batch normalization layer \cite{ioffe_batch-normalization_2015} and a dropout layer.
For the task-specific layers we experiment with dense layers, LSTMs \citep{hochreiter_long_1997}, GRUs \cite{cho_GRU_2014}, and temporal convolutional layers. Each task block has a different temporal receptive field and temporal dilation factors (\eg, using only every other image).
The output nodes of the conditional task blocks are divided into three equally sized groups. The directional command $c\in\{straight,\, left,\, right\}$ is used to switch between these groups of nodes. Only the active group is used to predict the target label. 
During training the gradients are only backpropagated through this group. 
We observed that training specialized submodules for each directional command leads to better performance compared to using the directional command as an additional input to the task networks. This observation agrees with \citet{codevilla_end--end_2017}. However, note that in contrast to their Conditional Imitation Learning approach, we use specialized task networks to predict conditional affordances rather than conditional steering commands.
Conditioning in the network has several advantages over conditioning in the controller, see the supplementary material for a discussion.

\textbf{Training.} As our approach decouples affordance prediction and control, we are able to train our perception stack independently from the chosen control algorithm. To train the affordance prediction network, we recorded a dataset with the CARLA simulator \cite{dosovitskiy_carla:_2017}.
CARLA provides two towns: Town 1 for training and Town 2 for testing.
For collecting training data in Town 1, we navigate through the city using our controller (described in section \ref{subsec:controller}) based on the ground truth affordances.
During data collection, the directional input at intersections is chosen at random.
All other traffic participants and all traffic lights are controlled by the in-game AI.
The full dataset contains approximately 240 thousand images and the corresponding ground truth labels for each of the six affordances. Five percent of the dataset is used for validation. The network is trained on mini-batches of size 32. For optimization we use Adam \cite{kingma_adam:_2014} with a base learning rate of 5e-5.
For classification we use the class-weighted categorical cross-entropy $H$ where the weights are chosen inversely proportional to the class label occurrence in the training set.
For regression, we use the mean average error $M\!AE$.
All six affordance loss functions are added to yield the loss function for the whole network:
\begin{align}
&\mathcal{L} = \sum_{j=1}^3 H_j + \sum_{k=1}^3 M\!AE_k.
\end{align}
Data augmentation is crucial to achieve generalization. Following \citep{dosovitskiy_carla:_2017}, we used the following augmentations: change in color, contrast, and brightness as well as Gaussian blur and salt-and-pepper noise.
Common techniques like vertical and horizontal flipping are not used, because in the driving context vertical flipping would correspond to switching from right- to left-hand traffic and horizontal flipping --- to driving upside down. 
In addition, we employ camera pose randomization. It is well known that for imitation learning utilizing a single fixed camera pose  leads to unstable behavior \citep{bojarski_end_2016,muller2018sim4cv} due to control errors and unexpected observations reinforcing each other.
Thus, we use a three-camera setup that records the same episode at $3$ different baselines ($-50$ cm, $0$ cm, $+50$ cm). In addition, each camera is randomly rotated around its vertical axis by up to $\pm \ang{15}$. This setup is only used for collecting the dataset. During testing, we use the input of a single centered camera.

During 10 \% of the training episodes, we additionally set the distance to the preceding vehicle $\ell$ to zero to provoke rear-end collisions with other cars which otherwise would not occur in our training set.
This way the network learns to correctly predict these situation as hazard stops.

\textbf{Hyperparameter search.} 
The task block hyperparameters entail the $type$ of layer (Dense, LSTM, GRU, Temporal Convolution), the number of $nodes$ within these layers, the dropout ratio $p$, the sequence length $seq$, and the dilation value $dil$.
The architecture of our network is optimized in a two-step procedure.
First, we randomly choose $type$ and sample values for $nodes, p, seq,$ and $dil$ from predefined value ranges (see supplementary material for the respective ranges).
We initialize the network with the same set of parameters for each task block and train it until the error on the validation set converges.
Second, the overall best performing task blocks are combined into a new model. The combined model is then trained jointly until the validation error converges. 
The hyperparameters of the best performing task blocks are shown in the supplementary material.

\subsection{Controller}\label{subsec:controller}
The control algorithm is split into longitudinal control (throttle, brake) and lateral control (steering) since it is possible to decouple these two tasks \citep{kosecka_a-comparative-study_1998}.

\textbf{Longitudinal control.} The longitudinal controller is subdivided into several states, see the supplementary material for an illustration. The states are (in ascending importance): \textit{cruising}, \textit{following}, \textit{over\_limit}, \textit{red\_light}, and \textit{hazard\_stop}. All states are mutually exclusive. For \textit{cruising} the target speed $v^*$ is the current speed limit. Given the current speed $v(t)$ we minimize the error term $e = v^*(t) - v(t)$ using a PID controller \citep{ang2005pid} for the throttle value. The brake is set to zero. The value of $v^*$ also depends on the current road type. If the agent receives a directional input to turn left or right, we reduce $v^*$ by $10$ km/h to enable smooth and safe navigation through turns. The state switches to \textit{following} if another car is less than $35$ m ahead.
We chose this static threshold, because we found that a dynamic threshold combined with the uncertainty of the network predictions leads to undesired state switching between following and cruising.
We will denote the distance to the front car as $\ell$ and the current speed limit as $v_{max}$. We use the optimal car-following model of \citet{chen_deepdriving:_2015} and minimize the error term
\begin{align}
& e(t) = v^*(t) - v(t) = v_{max}\left(1-exp\left(-\frac{c}{v_{max}}\ell(t)\right)-d\right) - v(t)
\end{align}
with a second PID controller. If the car is driving faster than the current speed limit, the state switches to \textit{over\_limit}, for a detected red light it switches to \textit{red\_light}, for a detected obstacle it switches to \textit{hazard\_stop}. For \textit{over\_limit}, \textit{red\_light}, and \textit{hazard\_stop}, the throttle and brake value depend on the vehicle speed, the prediction probability and the urgency to stop. The details as well as a description of the PID tuning procedure are provided in the supplementary material.

\textbf{Lateral control.} For lateral control we use the Stanley Controller (SC) \citep{thrun_stanley:_2006}. The SC uses two error metrics: the distance to centerline $d(t)$ and the relative angle $\psi(t)$. The control law to calculate the steering angle $\delta_{SC}(t)$ at the current vehicle speed $v(t)$ is given by
\begin{align}
&\delta_{SC}(t) = \psi(t) + \arctan \left(\frac{kd(t)}{v(t)}\right)
\end{align}
where $k$ is a gain parameter. 
To calculate the final steering output $\delta$, we add a damping term $\delta_{d}$ which reduces swaying on straight roads. $\delta_{d}(t)$ is calculated by using the current output of the control law $\delta_{SC}(t)$, the steering angle at the previous time step $\delta(t-1)$, and a positive damping constant $D$:
\begin{align}
&\delta(t) = \delta_{SC}(t) + \delta_{d}(t) \quad \text{where} \quad \delta_{d}(t) = - D * (\delta_{SC}(t) -\delta(t-1))
\end{align}

%% file: tex/04_results.tex
\section{Results}

We use the driving benchmark and evaluation protocol of \cite{dosovitskiy_carla:_2017} as the basis of our evaluation. The benchmark is composed of four different driving tasks with increasing difficulty: driving straight, driving through a single turn, navigating through the town taking several turns, and navigating through the town with additional dynamic obstacles (cars, pedestrians). The agent is initialized at a pre-defined point in the town and has to reach a goal under a task-specific time limit. The time limit equals the time needed to reach the goal when driving along the optimal path at $10$ km/h.

The simulation engine provides two different towns (Town 1, Town 2). Only Town 1 has been seen during training. The evaluation distinguishes between weather conditions that were seen during training and those that were not. For each combination of task, town, and weather condition, $25$ different episodes are evaluated. We compare our Conditional Affordance Learning (CAL) approach to Conditional Imitation Learning (CIL) \citep{codevilla_end--end_2017} as well as the Modular Pipeline (MP) and Reinforcement Learning (RL) approach presented in \citep{dosovitskiy_carla:_2017}. 
As stated in the original paper, the RL baseline cannot be claimed state-of-the-art as it still has room for improvement regarding its implementation.
Note that the baselines ignore all red lights and target a cruising speed of $20$ km/h. In contrast, our agent stops for red lights and recognizes speed signs. To provide a fair comparison to prior art, we limit our algorithm to a cruising speed of $20$ km/h for the quantitative evaluation against the baselines. 
We use the same model for all tasks and weather conditions without parameter fine-tuning.

\subsection{Goal-Directed Navigation}

Table \ref{tab:quantitative_analysis} shows our quantitative comparison with respect to the state-of-the-art on CARLA on the task of goal-directed navigation. 
Our agent outperforms the baselines in most of the tasks and under most conditions. As expected, performance decreases for all methods with increasing task difficulty. CAL particularly excels in generalizing to the new town, a scenario where most baselines did not perform well. For the new town under training weather conditions, the performance of the CAL agent is up to $68$ \% better than the best baseline.

CAL outperforms MP on almost all of the tasks except for the tasks in the training town under new weather conditions where MP performs slightly better. In almost all tasks and conditions, CAL shows better performance compared to CIL.
Prediction of the relative angle $\psi$ especially profits from the temporal structure of the CAL network as shown in the supplementary material. Also, the additional parameters of the controller (\eg, damping $D$) improve driving performance.
CAL is outperformed by CIL only on one task: driving straight under the training weather conditions in the test town. Our analysis revealed that this happens due to a systematic failure of the CAL perception stack in one specific weather condition where several trees cast shadows onto the street, affecting the prediction of the distance $d$ to centerline such that the agent occasionally drifts off the street.

\begin{table}
\caption{{\bf Quantitative Evaluation on Goal-Directed Navigation Tasks.} We show the percentage of successfully completed episodes per task and compare CAL to a Modular Pipeline (MP) \cite{dosovitskiy_carla:_2017}, Conditional Imitation Learning (CIL) \cite{codevilla_end--end_2017} and Reinforcement Learning (RL) \cite{dosovitskiy_carla:_2017}.}
\vspace{.2cm}
\label{tab:quantitative_analysis}
\small
\centering
\ra{1.1}
\resizebox{1.0\linewidth}{!}{
\begin{tabular}{lccccccccccccccccccc}
\toprule
             & \multicolumn{4}{c}{Training conditions} &  & \multicolumn{4}{c}{New weather}        &  & \multicolumn{4}{c}{New town}        &  & \multicolumn{4}{c}{\begin{tabular}[c]{@{}c@{}}New town and\\ new weather\end{tabular}} \\
Task         & MP  & CIL          & RL  & CAL           &  & MP           & CIL & RL & CAL           &  & MP & CIL         & RL & CAL          &  & MP                & CIL                & RL                & CAL                        \\ \midrule
Straight     & 98  & 95           & 89  & \textbf{100} &  & \textbf{100} & 98  & 86 & \textbf{100} &  & 92 & \textbf{97} & 74 & 93          &  & 50                & 80                 & 68                & \textbf{94}               \\
One turn     & 82  & 89           & 34  & \textbf{97}  &  & 95           & 90  & 16 & \textbf{96}  &  & 61 & 59          & 12 & \textbf{82} &  & 50                & 48                 & 20                & \textbf{72}               \\
Navigation   & 80  & 86           & 14  & \textbf{92}  &  & \textbf{94}  & 84  & 2  & 90           &  & 24 & 40          & 3  & \textbf{70} &  & 47                & 44                 & 6                 & \textbf{68}               \\
Nav. dynamic & 77  & \textbf{83}  & 7   & \textbf{83}  &  & \textbf{89}  & 82  & 2  & 82           &  & 24 & 38          & 2  & \textbf{64} &  & 44                & 42                 & 4                 & \textbf{64}               \\ \bottomrule
\end{tabular}%
}
\vspace{-0.5cm}
\end{table}

\subsection{Infraction Analysis}
Table \ref{tab:infraction_statistic} shows the average distances driven between two infractions 
during the hardest task of the benchmark: navigation with dynamic obstacles.
Under test conditions 
our approach is able to outperform the baselines in almost every measure. This indicates the strong generalization ability of our approach in comparison to the baselines. CAL performs particularly well in avoiding collisions with other vehicles and pedestrians.
In the new environment, our agent performs more than $10$ times better than the best baseline (RL). During testing, pedestrians are crossing the street more frequently than in the training scenario. The fact that our agent is still able to perform well demonstrates that the hazard stop network was able to predict this affordance robustly despite its sparsity.

Occasionally, the CAL agent drives on the sidewalk. This infraction occurs primarily during right turns.
Since rights turns in CARLA are generally sharper than left turns, the relative motion of pixels between consecutive frames is larger during a right turn. This may pose a bigger challenge to our perception network which is operating on sequential input.
Driving on the sidewalk sometimes also results in crashes with static objects, \eg, traffic lights. 
Note that the infraction count for static collisions is generally higher than that for driving on the sidewalk as the agent sometimes slides along a railing located next to the road, continuously increasing the infraction count.

\begin{table}
\small
\centering
\caption{{\bf Infraction Analysis}. We show the average distance (in km) driven between two infractions. '$>$' indicates that no infraction occurred over the whole course of testing. Note that the total driven distance depends on the amount and length of the episodes that an agent finishes.}
\vspace{.2cm}
\label{tab:infraction_statistic}
\ra{1.2}
\resizebox{1.0\linewidth}{!}{
\begin{tabular}{lccccccccccccccccccc}
\toprule
 & \multicolumn{4}{c}{Training conditions} &  & \multicolumn{4}{c}{New weather} &  & \multicolumn{4}{c}{New town} &  & \multicolumn{4}{c}{\begin{tabular}[c]{@{}c@{}}New town and\\ new weather\end{tabular}} \\
Task & MP & CIL & RL & CAL &  & MP & CIL & RL & CAL &  & MP & CIL & RL & CAL &  & MP & IL & RL & CAL \\ \midrule
Opposite lane & 10.2 & \textbf{33.4} & 0.18 & 6.7 &  & 16.1 & \textbf{57.3} & 0.09 & \textbf{\textgreater{}60} &  & 0.45 & 1.12 & 0.23 & \textbf{2.21} &  & 0.40 & 0.78 & 0.21 & \textbf{2.24} \\
Sidewalk & \textbf{18.3} & 12.9 & 0.75 & 6.1 &  & 24.2 & \textbf{\textgreater{}57} & 0.72 & 6.0 &  & 0.46 & 0.76 & 0.43 & \textbf{0.88} &  & 0.43 & 0.81 & 0.48 & \textbf{1.34} \\
Collision: Static & \textbf{10.0} & 5.38 & 0.42 & 2.5 &  & \textbf{16.1} & 4.05 & 0.24 & 6.0 &  & \textbf{0.44} & 0.40 & 0.23 & 0.36 &  & \textbf{0.45} & 0.28 & 0.25 & 0.31 \\
Collision: Car & \multicolumn{1}{l}{\textbf{16.4}} & \multicolumn{1}{l}{3.26} & \multicolumn{1}{l}{0.58} & \multicolumn{1}{l}{12.1} & \multicolumn{1}{l}{} & \multicolumn{1}{l}{20.2} & \multicolumn{1}{l}{1.86} & \multicolumn{1}{l}{0.85} & \multicolumn{1}{l}{\textbf{\textgreater{}60}} & \multicolumn{1}{l}{} & \multicolumn{1}{l}{0.51} & \multicolumn{1}{l}{0.59} & \multicolumn{1}{l}{0.41} & \multicolumn{1}{l}{\textbf{2.04}} & \multicolumn{1}{l}{} & \multicolumn{1}{l}{0.47} & \multicolumn{1}{l}{0.44} & \multicolumn{1}{l}{0.37} & \multicolumn{1}{l}{\textbf{1.68}} \\
Collision: Pedestrian & 18.9 & 6.35 & 17.8 & \textbf{30.3} &  & 20.4 & 11.2 & 20.6 & \textbf{\textgreater{}60} &  & 1.40 & 1.88 & 2.55 & \textbf{26.49} &  & 1.46 & 1.41 & 2.00 & \textbf{6.72} \\ \bottomrule
\end{tabular}
}
\end{table}

\subsection{Attention Analysis}

\begin{figure}
\centering    
  \begin{minipage}[t]{0.32\linewidth}
      \centering    
    \includegraphics[width=\textwidth]{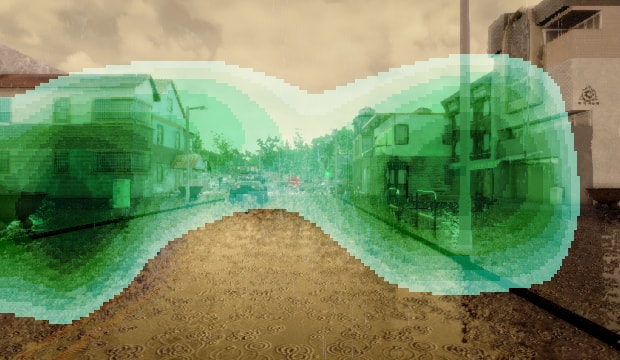}
  \end{minipage}
\hspace{0.05cm}
    \begin{minipage}[t]{0.32\linewidth}
      \centering    
    \includegraphics[width=\textwidth]{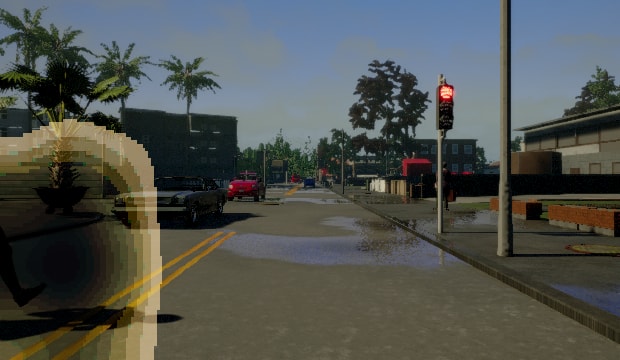}
  \end{minipage}
\hspace{0.05cm}
    \begin{minipage}[t]{0.32\linewidth}
      \centering    
    \includegraphics[width=\textwidth]{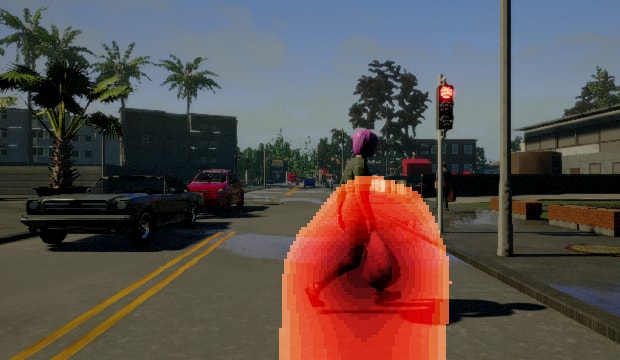}
  \end{minipage}
	\caption{{\bf Attention Analysis.} Left: The network focuses at the periphery for recognizing traffic lights. Middle: The network is anticipating a hazard stop by observing the shadow of a pedestrian. Right: The network focuses its attention on the pedestrian to correctly predict a hazard stop.}
	\label{fig:attention}
	\vspace{-0.3cm}
\end{figure}

In order to gain more insight into the decision making process of our perception stack we explore gradient-weighted class activation maps (Grad-CAMs) \cite{selvaraju_grad_2016} which expose the implicit attention of the CNN with respect to the input image.
Figure~\ref{fig:attention} shows the results for the affordances $hazard\_stop$ and $red\_light$. We encode the prediction probability for a particular class with different colors (green means low probability, red means high probability).
The attention given to a particular image region is encoded in the intensity channel.
The stronger the attention, the more intense the color. Regions with very low attention are shown transparent for clarity.
For imitation learning this attention analysis would be much less interpretable since predicting a single control output, e.g. throttle, entails the detection of speed limits, red lights, cars and pedestrians - all in one signal.

Figure~\ref{fig:attention} (left) shows the agent driving through town 1 with the attention map for $red\_light$ overlaid. Note that the network focuses its attention on the road outlines and structures on the roadside where traffic lights are expected to be located.
In Figure~\ref{fig:attention} (middle) the shadow of a pedestrian is visible at the bottom left of the image. Interestingly, the network anticipates an impending $hazard\_stop$ and focuses its attention to this region despite the fact that the pedestrian itself is not visible in the image.
Finally, in Figure~\ref{fig:attention} (right) we illustrate a hazard stop situation. The network focuses all its attention on the pedestrian's lower body. A possible explanation is that the network is able estimate the pedestrian's walking direction by observing its legs and feet.
Note that no pixel-wise supervision has been provided to the network during training. Instead, this behavior has been learned from binary image-level labels ($hazard\_stop=True/False$) alone.

%% file: tex/05_conclusion.tex
\section{Conclusion}\label{chap:conclusion} 
We have proposed Conditional Affordance Learning which combines the advantages of both modular pipelines and end-to-end approaches for navigating complex urban environments using high-level directional commands as input. We tested our approach extensively in simulation and demonstrated significant performance gains with respect to the state-of-the-art on CARLA. In the future, we plan to extend our work in several directions. Training on more diverse environments will likely improve the agent's performance in all domains. An extended study of network architectures can improve the quality of the predictions and therefore improve the driving capabilities of the agent. We also plan to extend the sensor setup, \eg, using a stereo camera to improve distance prediction. More sophisticated control algorithms like model predictive control may further improve the performance, in particular when training the parameters of the controller jointly with the perception stack. Finally, inspired by recent work \citep{muller_driving_2018}, we plan to transfer our results from simulation to the real world.

%% file: tex/A_00.tex
\vbox{%
\hsize\textwidth
\linewidth\hsize
\vskip 0.1in
\centering
{\LARGE\bf Supplementary Material:\\Conditional Affordance Learning\\for Driving in Urban Environments \par}
\vskip 0.2in
{\bf Axel Sauer\textsuperscript{\normalfont{1,2}} \quad
Nikolay Savinov\textsuperscript{\normalfont{1}} \quad
Andreas Geiger\textsuperscript{\normalfont{1,3}} }\\[2mm]
\textsuperscript{1} Computer Vision and Geometry Group, ETH Z{\"u}rich \\
\textsuperscript{2} Chair of Robotics Science and System Intelligence, Technical University of Munich \\
\textsuperscript{3} Autonomous Vision Group, MPI for Intelligent Systems and University of T{\"u}bingen
\vskip 0.2in
}

\begin{abstract}
	This {\bf supplementary document} provides further implementation details of our CAL agent in Section 1, a detailed description of our ground truth acquisition process in Section 2 and additional experiments in Section 3.\\The {\bf supplementary video} shows several navigation examples and visualizes the attention of our agent for different affordance indicators over time. The video is available at 
	\url{https://www.youtube.com/watch?v=UtUbpigMgr0}.
\end{abstract}

\setcounter{section}{0}

\input{tex/A_01_CAL_agent_details.tex}
\input{tex/A_02_GT.tex}
\input{tex/A_03_experiments.tex}

%% file: tex/A_01_CAL_agent_details.tex
\section{Implementation Details}\label{sec:cal_agent_details}
In this section, we list the value ranges and results of our hyperparameter search. We also provide additional details about our longitudinal control algorithm and the PID tuning procedure.

\subsection{Hyperparameter Search}\label{subsec:perception}
Table \ref{tab:hyper_params} shows each hyperparameter and its range of values for the random search described in the main paper. We initialize the network with randomly sampled parameters from the respective ranges. Table \ref{tab:best_task_blocks} shows the parameters of the best performing task blocks after optimization.

\begin{table}[h]
  \centering
  \ra{1.3}
  \caption{{\bf Hyperparameter Search.} Value ranges for each hyperparameter.}
  \vspace{1em}
  \label{tab:hyper_params}
  \resizebox{1\linewidth}{!}{
  \begin{tabular}{@{}c|ccccc@{}}
  \toprule
  Hyperparameter  & Layer type & Number of nodes & Dropout amount & Sequence length & Dilation value\\ 
  Acronym         & $type$ & $nodes$ & $p$ & $seq$ & $dil$ \\ \midrule
  Range of values &  \ra{1.2}\begin{tabular}[c]{@{}l@{}}\textit{Dense, GRU, LSTM,}\\ \textit{temporal convolution}\end{tabular} 
  & $[10,200]$ & $[0.25, 0.75]$ & $[1, 20]$ & $[1,3]$  \\ 
  \bottomrule
  \end{tabular}
  }
\end{table}

\begin{table}[h]
\centering
\footnotesize
\caption{{\bf Best task block parameters.} We optimized the layer $type$, the number of $nodes$, the dropout ratio $p$, the sequence length $seq$, and the dilation value $dil$.}
\label{tab:best_task_blocks}
\vspace{1em}
\ra{1.1}
\begin{tabular}{@{}lllcccc@{}}
\toprule
Task             &  &  $type$           & $nodes$ & $p$  & $seq$ & $dil$ \\ \midrule
Red light        &  & GRU 			    & 185     & 0.27 & 14    & 2        \\
Hazard stop      &  & Temp. convolution & 160     & 0.68 & 6     & 1        \\
Speed sign       &  & Dense 	   		& 160     & 0.55 & 1     & 1        \\
Vehicle distance &  & GRU			    & 160     & 0.38 & 11    & 1        \\
Relative angle   &  & Temp. convolution & 100     & 0.44 & 10    & 1        \\
Center distance  &  & Temp. convolution & 100     & 0.44 & 10    & 1        \\ \bottomrule
\end{tabular}
\end{table}

\clearpage

\subsection{Controller}\label{subsec:controller}
\begin{figure}[t]
\centering      
    \includegraphics[width=0.7\linewidth]{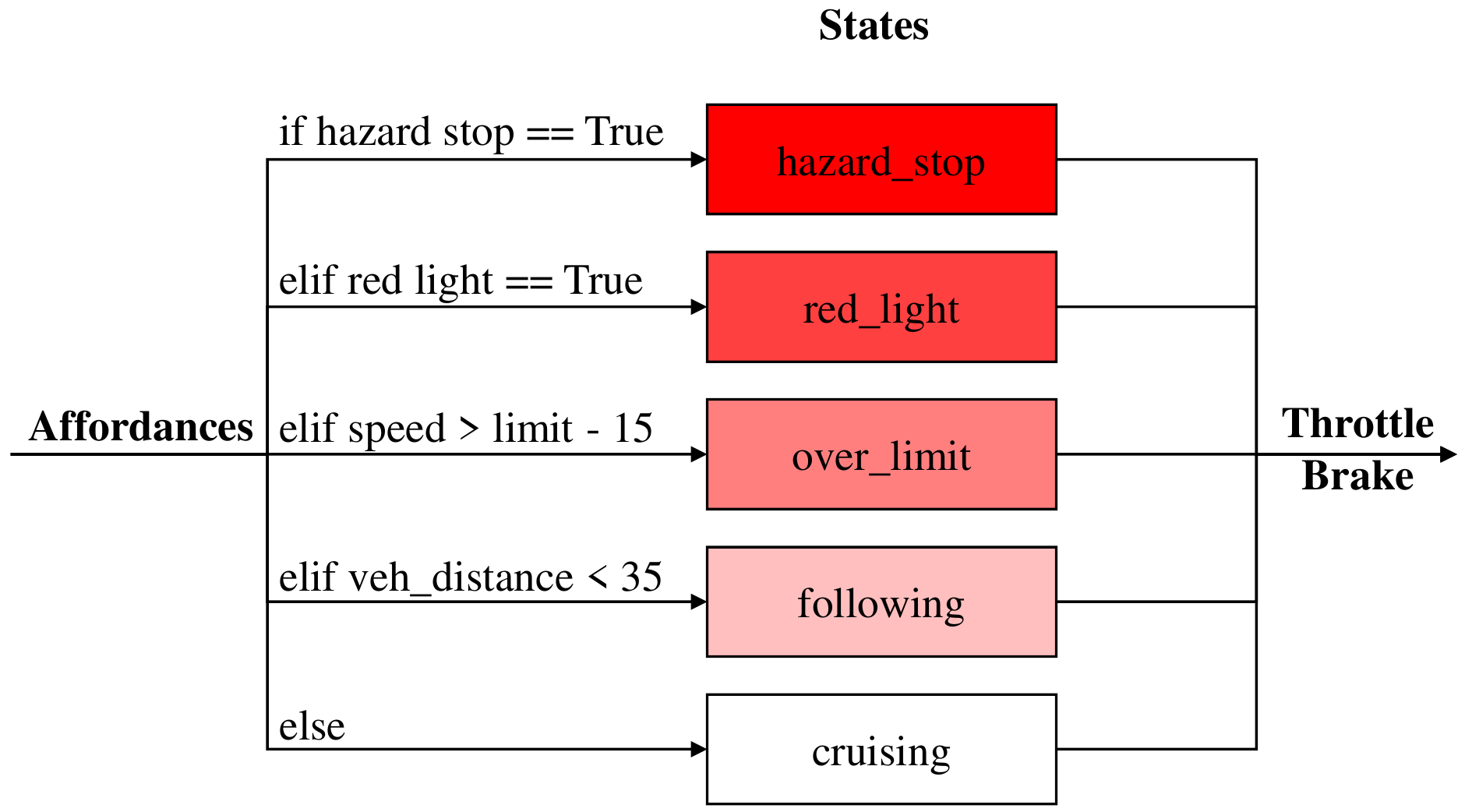}
	\caption{{\bf States of Longitudinal Controller.} The states are ordered in descending importance from top-to-bottom as indicated by the color intensity. All state are mutually exclusive.}
	\label{fig:controllers}
\end{figure}

\textbf{Longitudinal Control.} The states of the longitudinal controller are illustrated in Figure \ref{fig:controllers}. 

The throttle and brake values for the states \textit{over\_limit}, \textit{red\_light}, and \textit{hazard\_stop} are as follows:
\begin{itemize}
\item \textbf{over\_limit}:
This state is activated, if the agent is driving more than 15 km/h faster than the speed limit $v^*$. This situation typically occurs when entering a low-speed zone from a high-speed zone. To decelerate quickly, we set the throttle value to zero and calculate the brake value depending on the current speed $v(t)$:
\begin{align}
& brake = 0.3 \cdot \frac{v(t)}{v^*(t)}
\end{align}
As an example, driving with full speed (90 km/h) into a 30 km/h zone yields $brake = 0.9$, hence applying the brakes almost fully.

\item \textbf{red\_light}: 
If the prediction probability $P$ for the class \textit{red\_light}, i.e., the actual softmax output, is higher than a threshold $P_{rl}$, the controller switches to the state \textit{red\_light}. The throttle is set to zero and the brake is applied with:
\begin{align}
&brake = 0.2 \cdot \frac{v(t)}{30}
\end{align}
We empirically found that a threshold of $P_{rl}=0.9$ reduces false positives while still being able to reliable stop in front of red lights. Note that we use a smaller multiplier ($0.2$) compared to the $over\_limit$ state as red lights typically occur in 30 km/h zones. We consider the current speed $v(t)$ to gradually slow down the car in front of red lights.

\item \textbf{hazard\_stop}: This state is activated when an obstacle in front of the agent is detected, i.e., when $P(hazard\_stop) > P_{hs}$ where we empirically determined the threshold $P_{hs}$ as $P_{hs}=0.7$. Note the threshold is lower than that for the $red\_light$ state, since preventing crashes with road hazards is more critical for successful goal-directed navigation. When a $hazard\_stop$ has been detected, the throttle is set to zero and the brake is set to one.
\end{itemize}

\textbf{PID Controller Tuning.} The PID controllers used in the $cruising$ and $following$ state follow the standard PID control scheme \citep{ang2005pid}. The overall control function is given as follows
\begin{align}
& u(t) = K_pe(t) + K_i \int_0^t e(t')dt' + K_d\frac{de(t)}{dt},
\end{align}
where $K_p$, $K_i$ and $K_d$ are the proportional, integral and derivative coefficients. To be able to tune the coefficients of the two PID controllers, we implemented a visualization tool of speed, distance to the centerline, and other important signals. With this direct visual feedback, it is possible to use standard PID tuning methods. In this work, we leverage the method by Ziegler-Nichols \citep{ziegler1942optimum}. First, all coefficients are set to zero. Then, the proportional gain $K_{p}$ is increased until the output of the loop starts to oscillate. Given this ``ultimate gain'' $K_{u}$ and the oscillation period $T_{u}$, we set the coefficients to 
\begin{align}
&K_{p} = 0.6 K_{u}, \\
&K_{i} = T_{u}/2, \\
&K_{d} = T_{u}/8.
\end{align}
Using these values as a starting point, we empirically fine-tune the coefficients for optimal performance, with the goal of enabling fast but smooth reactions to disturbances.

\subsection{ Conditioning in the Network vs. Conditioning in the Controller}\label{subsec:conditioning}
Conditioning in the neural network is advantageous compared to directly conditioning the controllers. Conditioning the controllers would require the prediction of all affordances including the ones which are not relevant for the current navigation task at every point in time. 

Consider an intersection where the goal is to go straight. Even though the only relevant signal in terms of distance to the centerline is the distance to the centerline of the straight lane, the distance to the centerlines of the other lanes (left turning, right turning) must also be predicted in order to let the controller select the relevant signal based on the directional command. However, predicting the irrelevant affordances is very difficult as the required image features may not be present (when going straight, the turning lanes leave the field of view). Furthermore, the task is also highly ambiguous (once the intersection has been traversed via the straight lane, the distances to the centerline of the left and right turning lanes are not well defined). Thus, requiring the network to predict all affordances in every situation would introduce noise during training and hence lower the performance of the system. Besides, it would also increase runtime as more signals must be predicted at any point in time.

%% file: tex/A_02_GT.tex
\section{Ground Truth Acquisition}\label{sec:GT}
The API of CARLA supplies measurements about the agent (speed, acceleration, location, orientation) and about other objects in the scene (cars, pedestrians, traffic lights, and speed limit signs). These measurements include the current status of the traffic light (green, orange, red) and the type of the speed sign (30, 60, 90). Location and orientation are defined in a world coordinate system $\boldsymbol{c_{global}} = (x_g, y_g, z_g)^\intercal$. As these measurements do not directly express the affordances we want to learn, we implemented a procedure to convert them into the desired ground truth. 

\subsection{Observation Area}\label{subsec:observed_area}

We define a local coordinate system $\boldsymbol{c_{local}} = (x_l, y_l, z_l)^\intercal$ at the center of the front axle of the car with the x-axis corresponding to the car's lateral axis and the z-axis corresponding to the up vector. The agent's orientation $\psi$ and the agent's position $(x_{ego}, y_{ego})$ is supplied in $\boldsymbol{c_{global}}$.
Using this information, we convert the position of all other objects to $\boldsymbol{c_{local}}$. Next, we define the \textit{observation areas} as rectangles in the x-y plane of $\boldsymbol{c_{local}}$, see Figure 3 of the main paper for an illustration. If an object falls into an observation area it is considered ``detected''. 

\begin{table}
  \centering    
    \caption{{\bf Observation Areas.} The observation areas are rectangular boxes. This table lists the $(x,y)$ coordinates of the vertices $v$ of each observation area (provided in $\boldsymbol{c_{local}}$ in meters.)}
  \ra{1.5}
  \vspace{1em}
  \begin{tabular}[b]{@{}clcccc@{}}
  \toprule
  Area  & Detection of                                                                        
  & $v_1$ & $v_2$ & $v_3$ & $v_4$\\ \midrule
  $A1$  & \ra{1.2}\begin{tabular}[c]{@{}l@{}}\textit{red\_light}\\ \textit{speed\_sign}\end{tabular}
  & (7.4, -0.8) & (7.4, -5.8) & (14.0, -0.8) & (14.0, -5.8) \\ 
  $A2$  & \textit{hazard\_stop}              
  & (0.0, 2.0) & (0.0, -2.0) & (8.2,2.0) & (8.2,-2.0)\\ 
  $A3$  & \textit{distance\_to\_vehicle}      
  & (0.0, 1.6) & (0.0, -1.6) & (50.0, 1.6) & (50.0, -1.6) \\ 
  \bottomrule
  \end{tabular}
  \label{fig:observed_area}
  \vspace{-1em}
\end{table}

The length, width, position, and orientation of the observation areas are chosen based on the respective affordances. Thus, the observation area for $red\_light$ and for $speed\_sign$ is located on the right side of the agent as their respective signals are located on the right sidewalk. The observation area for $hazard\_stop$ is directly in front of the car and very short, in order to only initiate a hazard stop if an accident could not be avoided otherwise. The observation area for $vehicle\_distance$ is in front of the car and has a length of 50 m. If another vehicle is located within this area, the distance of the closest vehicle to the agent is measured. If there is no car in front, the default value for $vehicle\_distance$ (50 m) is used. Table \ref{fig:observed_area} lists the coordinates for each observation area.

\subsection{Directional Input}\label{subsec:highlevel_planner}
CARLA provides a high-level topological planner based on the A* algorithm. It takes the agents position and the coordinates of the destination and calculates a list of commands. This ``plan'' advises the agent to turn left, right or to keep straight at intersections. 

%% file: tex/A_03_experiments.tex
\section{Additional Experiments}\label{sec:experiments}

In this section, we provide additional experiments. First, we compare temporal to non-temporal task blocks to assess the influence of the additional temporal information provided by video data. Second, we provide a qualitative evaluation of our agent's driving behavior.

\subsection{Comparison of Temporal and Non-temporal Task Blocks}\label{subsec:comparison_non_temporal}
Table \ref{tab:comparison_non_temporal} shows the best performing task blocks for each task. By using a temporal task block, all the classification and regression results improve. This demonstrates that each task profits from the additional temporal information. 

\begin{table}
\caption{{\bf Comparison of Temporal and Non-temporal Task Blocks}. The last column shows the relative change in performance. The higher the $\overline{IoU}$ and the lower the $M\!AE$ the better.}
\label{tab:comparison_non_temporal}
\vspace{1em}
\ra{1.2}
\centering
\resizebox{1.0\linewidth}{!}{
\begin{tabular}{@{}C{3cm}C{1.5cm}C{2.5cm}C{2.5cm}C{3cm}@{}}
\toprule
 &&   \multicolumn{2}{c}{Best performing task blocks} \\ \cmidrule(lr{1em}){3-4} 
Task & Metric& non-temporal & temporal & relative Change\\ \midrule
Hazard stop      & $\overline{IoU}$ & 84.96 \% & 87.41 \% & + 2.88 \% \\ 
Speed sign       & $\overline{IoU}$ & 91.95 \% & 92.71 \% & + 0.83 \% \\ 
Red light        & $\overline{IoU}$ & 92.41 \% & 93.95 \% & + 1.67 \% \\ 
Relative angle   & MAE & 0.00797 & 0.00207 & - 74.03 \% \\ 
Center distance  & MAE & 0.09642 & 0.08465 & - 12.21 \% \\ 
Vehicle distance & MAE & 0.04497 & 0.03289 & - 26.86 \% \\ \bottomrule
\end{tabular}
}
\end{table}

The biggest relative improvement can be seen for the \textit{relative\_angle} task block. The error of the temporal task block is almost four times lower than the non-temporal task block. This suggests that this task profits more from the temporal context than other tasks. 
The smallest improvement is achieved for the \textit{speed\_sign} task. 
To keep  computation time low during training and inference, we therefore use the non-temporal task block for this task in our final model.

In addition, we empirically observed that there is no dominating temporal layer in terms of performance. LSTMs, GRUs and temporal convolution layers perform very similar.

\subsection{Driving Behaviour}\label{subsec:driving_behaviour}
The experiments in the main text examined whether the goal was reached and if there were any rule violations. This section focuses on qualitative driving experience, i.e., how the driving would be perceived by a passenger. The evaluation is done for the task ``navigation without dynamic objects'' to evaluate the general driving behavior without distorting the results by the challenges of stopping for cars or pedestrians. 

We use the following metrics for evaluation:
\begin{itemize}
\item \textbf{Centerline distance:} Staying in the middle of the road is the main task of every lane keeping system. We evaluate the ability of the algorithm to minimize the deviation from the centerline. The reported result is the median over all episodes. The median is more descriptive for the qualitative driving experience than the mean value since failed episodes during which an agent drifts off the road produce large outliers.
\item \textbf{Jerk} is the rate of change of acceleration. The jerk can be felt during driving in the form of a jolt or sudden shock. It is commonly used to quantify ride comfort \citep{huang_fundamental_2004}. A smoother ride results in lower fuel consumption increased passenger comfort and more safety. A way to quantify jerk is to compare the root mean square (RMS) jerk \citep{hrovat_1981_optimum}. The analysis further distinguishes between longitudinal and lateral jerk with lateral jerk separately evaluated for straight roads and in turns. 
\end{itemize}
   
Table \ref{tab:driving_behavior_benchmark} reports our results.
In contrast to the previous evaluations, the results are not reported depending on the weathers conditions or the environments.
The results are very similar under all conditions for all agents. The CAL agent is able to achieve the best performance on all four metrics.

\subsubsection*{Distance to Centerline}
All agents perform on a similar level and are able to keep the distance to the centerline low. The exception is the RL approach. When driving on a straight road, the RL agent regularly starts swaying from side to side over the entire lane, resulting in the high value.

\subsubsection*{Longitudinal Jerk}
The CAL agent performs best, followed by CIL and RL. The control parameters of the CAL agent are freely adjustable which allows to accelerate and decelerate smoothly as well as driving at a constant pace.
The RL agent is only able to set the throttle value to either 0 or 1. This results in a sudden jerk every time the agent utilizes the throttle.

\subsubsection*{Lateral Jerk while Driving Straight}
On straight roads, both the CAL and the CIL agent perform similarly. When driving straight, the RL agent often outputs a steering value of 0. This leads to the agent drifting off the road. When correcting against the drift, the RL agent steers abruptly, resulting in large jerk values.

\subsubsection*{Lateral Jerk while Turning}
The CAL agent performs exceptionally well. There are several reasons for this. First, the agent is slowing down when approaching a turn. Second, our agent turns smoothly without abrupt changes in steering. Third, the jerk peaks are generally lower than for the other approaches. Despite this good performance, the transition from turns to straight roads leaves room for improvement. Changes in the directional switch result in sudden jumps in the prediction of the relative angle in some cases, resulting in slight short-timed jerk. The CIL agent is not as good as the CAL agent, but it is generally able to drive through turns smoothly. RL, in contrast, conducts strong and abrupt steering movements, resulting in a higher jerk value.

\begin{table}
\small
\centering
\caption{{\bf Driving Behaviour.} Qualitative evaluation of the general driving performance, the lower a metric the better.}
\vspace{1em}
\label{tab:driving_behavior_benchmark}
\ra{1.2}
\begin{tabular}{lcccccc}
\toprule
Metrics                        & Unit      &  & CIL   & RL    & CAL             \\ \midrule
Distance to centerline         & $[m]$     &  & 0.390 & 0.755 & \textbf{0.334}   \\
Longitudinal jerk              & $[m/s^3]$ &  & 0.449 & 1.368 & \textbf{0.333}    \\
Lateral jerk  driving straight & $[m/s^3]$ &  & 0.084 & 0.336 & \textbf{0.052}    \\
Lateral jerk driving turns     & $[m/s^3]$  &  & 0.242 & 0.548 & \textbf{0.065}    \\ \bottomrule
\end{tabular}
\end{table}